\documentclass{article}

\PassOptionsToPackage{numbers,sort&compress}{natbib}

\usepackage{amsmath}
\usepackage{amssymb}
\usepackage{amsfonts}
\usepackage{mathtools}
\usepackage{enumitem}
\usepackage{booktabs}
\usepackage{graphicx}
\usepackage{xcolor}
\usepackage{tabularx}

\usepackage[dblblindworkshop, final]{neurips_2026}
\workshoptitle{AI for Drug Discovery: Bridging the Translation Gap}

\usepackage[utf8]{inputenc} 
\usepackage[T1]{fontenc}    
\usepackage{hyperref}       
\usepackage{url}            
\usepackage{booktabs}       
\usepackage{amsfonts}       
\usepackage{nicefrac}       
\usepackage{microtype}      
\usepackage{xcolor}         
\title{VCLMU: Mechanism-Centric Virtual Cell World Modeling for Perturbation Response}

\author{\begin{tabular}{ccc} Yuwei Miao$^{1}$\thanks{Equal contribution} & Azim Dehghani Amirabad$^{1}$\footnotemark[1] & Scott Oloff$^{1}$ \\[0.4em] Junzhou Huang$^{2}$\thanks{Corresponding authors: Junzhou Huang (jzhuang@exchange.uta.edu), Tianyu Cui (TCui8@its.jnj.com), and Rui Liao (RLiao2@its.jnj.com)} & Tianyu Cui$^{1}$\footnotemark[2] & Rui Liao$^{1}$\footnotemark[2] \end{tabular} \\[0.8em] $^{1}$Johnson \& Johnson Innovative Medicine \qquad $^{2}$University of Texas at Arlington}

\begin{document}

\maketitle

\begin{abstract}
Predicting cellular responses to genetic perturbations is a central capability for virtual cells and a key step toward computational modeling of biological interventions. Most existing models directly map an unperturbed molecular profile and perturbation to the resulting observation without explicitly representing the latent cellular transition induced by the intervention. We introduce a \textbf{mechanism-centric virtual cell world model} that represents cellular state as a set of Latent Mechanism Units (LMUs) and treats genetic perturbations as actions on these latent states. Each LMU combines a reusable identity grounded in multimodal biological evidence with an observation-specific state, allowing a perturbation to induce mechanism-specific stochastic transitions before decoding the resulting transcriptional response. We train VCLMU through two-stage pretraining, first on around 200K pseudo-bulk perturbation profiles and then on gene-aligned single-cell perturbation data. Across six perturbation-disjoint benchmarks, VCLMU consistently improves perturbation-specific response recovery over strong baselines while maintaining competitive global response accuracy. We further analyze learned LMUs through enrichment between perturbation responses and LMU gene sets and show that they capture structured biological response programs. These results support mechanism-level latent state transition as a useful formulation for virtual cell models that aim to predict and interpret cellular responses to biological interventions.

\end{abstract}

\begin{figure}[t]
    \centering
    \includegraphics[width=1\linewidth]{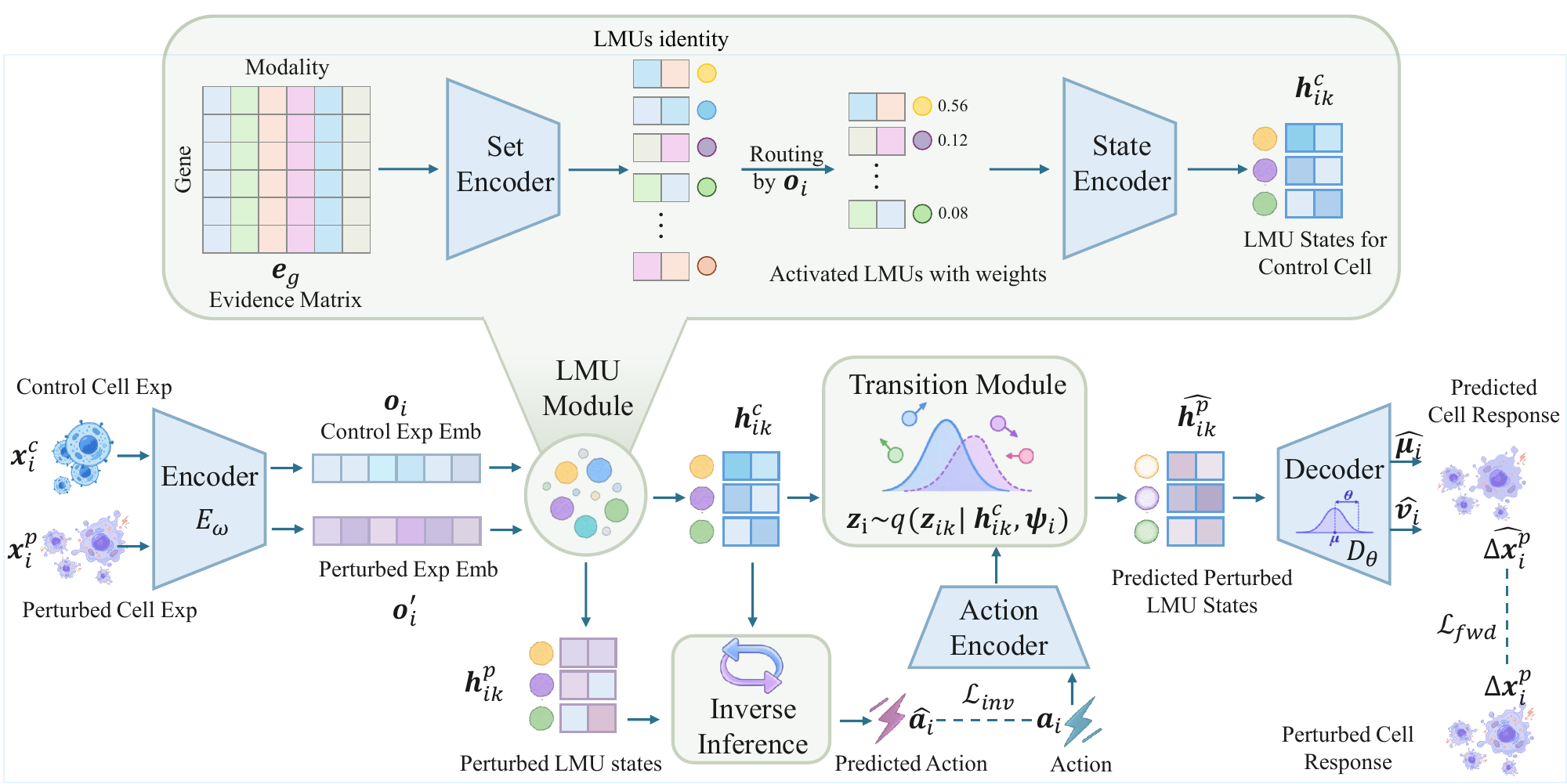}
    \caption{
    \textbf{Overview of the mechanism-centric virtual cell world model.}
    Multimodal biological evidence is composed into reusable Latent Mechanism Unit (LMU) identities. A control profile $\mathbf{x}_i^{c}$ is encoded and routed to a sparse active LMU set with control mechanism states $\mathbf{H}_i^{c}$, while the perturbation action $a_i$ is encoded as $\boldsymbol{\psi}_i$. Each active LMU undergoes an action-conditioned stochastic transition to produce predicted perturbed mechanism states, which are decoded into the response distribution $p(\Delta\mathbf{x}_i\mid\mathbf{x}_i^{c},a_i)$ and the predicted profile $\widehat{\mathbf{x}}_i^{p}=\mathbf{x}_i^{c}+\widehat{\boldsymbol{\mu}}_i$. During training, the observed perturbed profile is encoded through the same active LMUs to provide transition targets and support inverse prediction of the perturbation, without entering the forward prediction path.
    }
    \label{fig:framework}
\end{figure}

\begin{figure}[t!]
    \centering
    \includegraphics[width=0.8\linewidth]{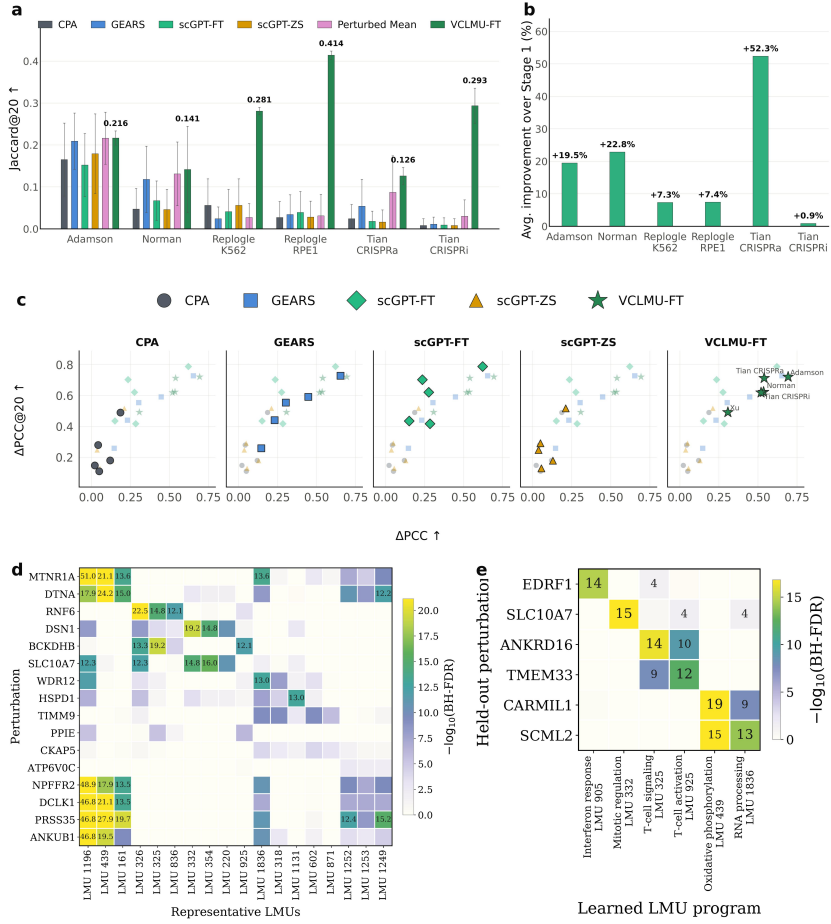}
    \caption{
    \textbf{Perturbation prediction and LMU interpretability.}
    \textbf{(a)} Jaccard@20 across six evaluation datasets for VCLMU-FT and baselines.
    \textbf{(b)} Stage~1 versus Stage~2 pretraining under the same downstream fine-tuning protocol; annotations indicate the Jaccard@20 gain.
    \textbf{(c)} $\Delta$PCC versus $\Delta$PCC@20, showing global and DE-focused Pearson correlation jointly.
    \textbf{(d)} Enrichment of perturbation-specific response genes in learned LMU candidate gene sets across held-out perturbations after excluding ubiquitous response genes and subtracting the shared response; color indicates $-\log_{10}(\mathrm{BH\text{-}FDR})$.
    \textbf{(e)} Six representative significant associations between perturbations and LMUs whose overlapping response genes form biologically coherent programs spanning interferon response, mitotic regulation, T-cell signaling, T-cell activation, oxidative phosphorylation, and RNA processing; numbers indicate overlapping response genes.
    }
    \label{fig:main_results}
\end{figure}
\section{Introduction}

Predicting how cells respond to genetic perturbations is a foundational capability for \emph{virtual cell} models and an important step toward computational evaluation of biological interventions~\citep{bunne2024virtualcell}. High-throughput single-cell perturbation screens~\citep{dixit2016perturbseq,adamson2016,norman2019,replogle2022}, now consolidated into harmonized resources~\citep{peidli2024scperturb}, provide increasingly large datasets for learning intervention-response relationships. A rapid progression of models has followed, including scGen~\citep{lotfollahi2019scgen}, CPA~\citep{lotfollahi2023cpa}, and GEARS~\citep{roohani2024gears}; pretrained transcriptome models~\citep{theodoris2023geneformer,cui2024scgpt,hao2024scfoundation}; and large-scale virtual cell models trained across diverse perturbations and cellular contexts~\citep{adduri2025state,wang2026xcell}, alongside standardized benchmarks~\citep{roohani2025vcc}.

Most existing perturbation predictors remain primarily \emph{observation-centric}: they map an unperturbed profile and intervention to the resulting transcriptional response while leaving the latent cellular transition induced by that intervention implicit. Scaling this map has so far yielded limited returns. A recent systematic evaluation finds that deep perturbation predictors still do not outperform simple linear or condition-agnostic baselines~\citep{ahlmanneltze2025linear}, suggesting that improved modeling of intervention-induced state changes may be as important as additional model capacity. Latent-variable models such as CPA~\citep{lotfollahi2023cpa} and SAMS-VAE~\citep{bereket2023samsvae} introduce structured latent effects, but a perturbation still enters primarily as an additive or conditioning term rather than as an action that transforms cellular state.

We instead adopt a single-step world-model formulation that encodes an observation into latent state, transitions that state under a perturbation, and decodes the resulting response~\citep{ha2018worldmodels,hafner2019planet,hafner2020dreamer,hafner2025dreamerv3}. Because concurrent cellular programs can respond differently to the same intervention, we factorize state into mechanism-specific units, following the general principle of object-centric world models~\citep{locatello2020slotattention,daniel2024ddlp,villar2025playslot,daniel2026lpwm}.

We therefore introduce VCLMU, which represents a cell as a sparse set of Latent Mechanism Units (LMUs) selected from a bank of $K$ units by a control-conditioned top-$K_a$ router (Fig.~\ref{fig:framework}). Each LMU separates a reusable identity from an observation-specific state: the identity represents which mechanism the unit captures and combines a learnable query with a set-encoder summary of a periodically refreshed candidate gene set grounded in multimodal gene evidence, while the state reflects the current activity of that mechanism from the expression of the same genes. Perturbations are encoded from the target gene and its regulatory neighborhood and applied independently to each active LMU, allowing heterogeneous responses across mechanisms. State transitions are residual and stochastic, with each unit drawing an action-conditioned Gaussian latent variable to represent variability not determined by the control state and perturbation. The transitioned states are then decoded into the perturbed expression profile. We explicitly supervise these latent transitions rather than only the final reconstruction: the observed perturbed profile is encoded through the same control-selected LMUs, and predicted state changes are aligned with the observed changes using stop-gradient targets, while the target branch is constrained to reconstruct the observed profile. The forward dynamics depend only on the control state and perturbation, preventing information leakage from the observed outcome.

Our contributions are (i) a formulation of perturbation response as action-conditioned latent-state transition in addition to direct observation regression; (ii) LMUs, a factorized cellular state whose units carry separate stochastic dynamics under a shared action; and (iii) supervision of the latent transition itself, with stop-gradient targets and observation grounding that keep the learned target from collapsing, together with a two-stage curriculum that pretrains on pseudo-bulk perturbation profiles and continues on gene-aligned single-cell data. Across six perturbation benchmarks, VCLMU improves recovery of top response genes over strong baselines without sacrificing global response accuracy, and its learned units align with coherent, largely disjoint transcriptional programs.

\section{Method}
\label{sec:method}

\paragraph{Problem formulation.}
Let $\mathbf{x}_i^{c},\mathbf{x}_i^{p}\in\mathbb{R}^{G}$ denote a control reference and an associated perturbed expression profile, and let $a_i$ index the perturbed gene. We predict the response $\Delta\mathbf{x}_i=\mathbf{x}_i^{p}-\mathbf{x}_i^{c}$ from $(\mathbf{x}_i^{c},a_i)$. Each gene $g$ carries a multimodal evidence embedding $\mathbf{e}_g\in\mathbb{R}^{d}$ built from functional annotation, pathway, interaction network, protein sequence, and tissue-expression resources, together with a gene-regulatory-network (GRN) neighborhood~\citep{kamimoto2023celloracle}. These features provide biological evidence for gene representation, while the response model itself is learned end-to-end. Details of the evidence representations are provided in Appendix~\ref{app:evidence}. Rather than fitting $(\mathbf{x}^{c},a)\mapsto\Delta\mathbf{x}$ directly, we factor the map through a latent state in the manner of a world model~\citep{ha2018worldmodels,hafner2020dreamer}: encode the control cell into a structured latent state, transition that state under the perturbation, and decode the transitioned state into the transcriptional response.

\paragraph{Cellular state as a set of Latent Mechanism Units.}
A cell is represented by a set of $K$ Latent Mechanism Units (LMUs), of which only a sparse subset $\mathcal{A}_i$ of size $K_a$ is active per cell, in the spirit of slot-based object-centric encoders~\citep{locatello2020slotattention} and sparsely-gated expert routing~\citep{shazeer2017moe}. Each LMU factorizes into an identity and a state, which differ in scope. The identity $\boldsymbol{\ell}_k$ is shared across all cells and grounded in biological evidence: LMU $k$ owns a small, periodically refreshed set of candidate genes $\mathcal{C}_k$, and its identity blends a free learnable query $\mathbf{q}_k$ with a permutation-invariant summary $\operatorname{SetEnc}$~\citep{lee2019settransformer} of that set's evidence embeddings,
\begin{equation}
\boldsymbol{\ell}_k
=
\operatorname{LayerNorm}\!\left(
\alpha\mathbf{q}_k+
(1-\alpha)\operatorname{SetEnc}\{\mathbf{e}_g:g\in\mathcal{C}_k\}
\right),
\label{eq:lmu_identity}
\end{equation}
where $\alpha\in(0,1)$ is learned. The free query allows each LMU to adapt to programs supported by the data, while the evidence term anchors its identity to biologically informed gene representations, yielding mechanism-level representations that are both biologically coherent and interpretable (Fig.~\ref{fig:main_results}d,e). The state $\mathbf{h}_{ik}$ is instead specific to a single observation: given a control profile $\mathbf{x}_i^{c}$, the observation encoder produces $\mathbf{o}_i^{c}=f_{\mathrm{obs}}(\mathbf{x}_i^{c})$, which routes to the active set and provides global context for LMU state construction,
\begin{equation}
\mathcal{A}_i=\operatorname{Route}(\mathbf{o}_i^{c}),
\qquad
\mathbf{h}_{ik}^{c}
=
f_{\mathrm{state}}\!\left(
\boldsymbol{\ell}_k,
\{(\mathbf{e}_g,x_{ig}^{c})\}_{g\in\mathcal{C}_k},
\mathbf{o}_i^{c}
\right),
\quad k\in\mathcal{A}_i.
\label{eq:lmu_state}
\end{equation}
Identity therefore answers which mechanism a unit is, and state answers how active it currently is. Routing depends only on the control profile, so the active set is a property of the starting cell rather than of the outcome. We write $\mathbf{h}_{ik}^{c}=\mathbf{h}_{ik}(\mathbf{x}_i^{c})$ and stack the active states of a sample into $\mathbf{H}_i^{c}\in\mathbb{R}^{K_a\times d}$. During training we additionally encode the observed perturbed profile as $\mathbf{h}_{ik}^{p}=\mathbf{h}_{ik}(\mathbf{x}_i^{p})$, stacked into $\mathbf{H}_i^{p}$, using the same indices $\mathcal{A}_i$: reusing the control-selected set puts the two representations in LMU-wise correspondence, so ``how much did mechanism $k$ move'' is well defined. Routing, evidence construction, and candidate refresh are detailed in Appendix~\ref{app:method_details}.

\paragraph{Perturbation as an action on mechanism states.}
A perturbation is encoded once, into an action embedding $\boldsymbol{\psi}_i=f_{\mathrm{act}}(a_i,\{\mathbf{e}_g\},\mathrm{GRN})$ that summarizes the target gene together with its GRN neighborhood, and is then applied separately to each active LMU. Because every unit receives the same action but occupies a different state, one perturbation can strongly move some mechanisms and leave others untouched, a behavior that is difficult to represent with a single global latent state. Transitions are residual and stochastic, with a per-unit latent $\mathbf{z}_{ik}$ drawn from a learned conditional distribution $q(\mathbf{z}_{ik}\mid\mathbf{h}_{ik}^{c},\boldsymbol{\psi}_i)$ and regularized toward the prior $p(\mathbf{z}_{ik})=\mathcal{N}(\mathbf{0},\mathbf{I})$ through a KL penalty~\citep{kingma2014vae}, allowing the transition to represent variability not determined by the control state and the action:
\begin{equation}
\widehat{\mathbf{h}}_{ik}^{p}
=
\mathbf{h}_{ik}^{c}
+
f_{\mathrm{dyn}}\!\left(
\mathbf{h}_{ik}^{c},\boldsymbol{\psi}_i,\mathbf{z}_{ik}
\right),
\qquad
\mathbf{z}_{ik}\sim q\!\left(\mathbf{z}_{ik}\mid\mathbf{h}_{ik}^{c},\boldsymbol{\psi}_i\right).
\label{eq:lmu_dynamics}
\end{equation}
The residual form means the model predicts a change to each mechanism rather than re-deriving the cell from scratch. Critically, the dynamics see only the control state and the action, so the observed outcome cannot leak into the forward path. Stacking $\widehat{\mathbf{h}}_{ik}^{p}$ into $\widehat{\mathbf{H}}_i^{p}$ as before, we supervise the dynamics by matching the predicted mechanism change to the observed one,
\begin{equation}
\mathcal{L}_{\mathrm{dyn}}
\!=\!
\operatorname{SmoothL1}\!\left(
\widehat{\mathbf{H}}_i^{p}-\mathbf{H}_i^{c},
\operatorname{sg}\!\left[\mathbf{H}_i^{p}-\mathbf{H}_i^{c}\right]
\right),
\label{eq:lmu_matching}
\end{equation}
where $\operatorname{sg}$ is a stop-gradient~\citep{oord2017vqvae}: the encoder supplies the target but is not trained to make that target easy to hit.

\paragraph{Decoding the predicted state.}
The transitioned units are first mixed by a lightweight interaction layer, since mechanisms are not independent, and then read out by a cross-attention decoder~\citep{vaswani2017attention} in which each gene attends to the predicted mechanism states using its own evidence embedding as the query. This produces a per-gene mean and variance,
\begin{equation}
(\widehat{\boldsymbol{\mu}}_i,\widehat{\mathbf{v}}_i)
=
f_{\mathrm{dec}}\!\left(
\{\widehat{\mathbf{h}}_{ik}^{p}\}_{k\in\mathcal{A}_i},
\{\mathbf{e}_g\}_{g=1}^{G}
\right),
\label{eq:lmu_decoder}
\end{equation}
defining a factorized Gaussian $p(\Delta\mathbf{x}_i\mid\mathbf{x}_i^{c},a_i)$ over the response. We identify the predicted response with its mean, $\widehat{\Delta\mathbf{x}}_i=\widehat{\boldsymbol{\mu}}_i$, giving the point prediction $\widehat{\mathbf{x}}_i^{p}=\mathbf{x}_i^{c}+\widehat{\Delta\mathbf{x}}_i$. Using gene evidence both to define LMU identities and to query the predicted mechanism states keeps the latent and observation spaces grounded in the same biological vocabulary.

\paragraph{Learning objective.}
Supervising only the decoded profile would leave the latent transition underdetermined, since many latent trajectories explain the same output. The full objective therefore combines the response likelihood with mechanism-level supervision and a small set of regularizers,
\begin{equation}
\mathcal{L}
=
\mathcal{L}_{\mathrm{fwd}}
+\lambda_{\mathrm{dyn}}\mathcal{L}_{\mathrm{dyn}}
+\lambda_{\mathrm{KL}}\mathcal{L}_{\mathrm{KL}}
+\lambda_{\mathrm{obs}}\mathcal{L}_{\mathrm{obs}}
+\lambda_{\mathrm{inv}}\mathcal{L}_{\mathrm{inv}}
+\mathcal{R}_{\mathrm{LMU}},
\label{eq:lmu_objective}
\end{equation}
where $\mathcal{L}_{\mathrm{fwd}}$ is the Gaussian negative log-likelihood of Eq.~\eqref{eq:lmu_decoder}, $\mathcal{L}_{\mathrm{dyn}}$ is the dynamics-alignment loss of Eq.~\eqref{eq:lmu_matching}, and $\mathcal{L}_{\mathrm{KL}}$ regularizes the transition distribution of Eq.~\eqref{eq:lmu_dynamics}. The remaining terms guard against degenerate solutions: $\mathcal{L}_{\mathrm{obs}}$ requires the observed mechanism states to reconstruct the observed profile, preventing the transition target from collapsing to an arbitrary code; $\mathcal{L}_{\mathrm{inv}}$ asks an inverse model to recover the perturbation from the control and perturbed mechanism states, their difference, and their elementwise interaction, encouraging the encoded state changes to retain perturbation-specific information; and $\mathcal{R}_{\mathrm{LMU}}$ collects weak identity-anchor, diversity, and activity-entropy terms that keep LMUs distinct and used. We pretrain on perturbation-disjoint pseudo-bulk profiles and then continue on gene-aligned single-cell data, with dynamics alignment and target-state grounding disabled during Stage~2 (Appendix~\ref{app:method_details}).

\section{Results}

\paragraph{Setup.}
We evaluate on held-out single-gene perturbations under perturbation-disjoint downstream splits on widely used genetic perturbation datasets, so no perturbation in the downstream fine-tuning set appears at test time (see more details in Appendix~\ref{sec:exp}). VCLMU-FT denotes the Stage-2 pretrained model further fine-tuned on the downstream training split of each evaluation dataset. We report Jaccard@20 together with Pearson correlation on the full perturbation response ($\Delta$PCC) and the top responsive genes ($\Delta$PCC@20).

\paragraph{Held-out perturbation prediction.}
VCLMU-FT attains the best Jaccard@20 on all six datasets (Fig.~\ref{fig:main_results}a). The margin is small on Adamson~\citep{adamson2016} and Norman~\citep{norman2019}, where the condition-agnostic Perturbed Mean is already competitive, consistent with reports that simple baselines remain hard to beat~\citep{ahlmanneltze2025linear}. The gains are largest on Replogle K562 and RPE1~\citep{replogle2022} and Tian CRISPRi~\citep{tian2021}, where baselines recover few true response genes. Using the Stage~2 pretrained checkpoint for downstream fine-tuning improves Jaccard@20 on five of six datasets, with the largest gains on Adamson, Norman, and Tian CRISPRa, while preserving performance on Replogle RPE1 (Fig.~\ref{fig:main_results}b). This indicates that pseudo-bulk pretraining captures transferable perturbation structure that subsequent single-cell pretraining further sharpens. VCLMU also occupies the upper-right region of the $\Delta$PCC/$\Delta$PCC@20 plane (Fig.~\ref{fig:main_results}c): accuracy on responsive genes does not come at the cost of the global response.

\paragraph{LMU candidate gene sets are associated with distinct perturbation-response programs.}
To assess the biological coherence of learned LMUs, we analyzed enrichment between perturbation-specific response genes and LMU candidate gene sets. Among 1{,}101 held-out perturbations with at least ten observations, after excluding ubiquitous response genes and subtracting the shared response, 86.6\% are significantly associated with at least one LMU (BH-FDR $<0.05$)~\citep{benjamini1995fdr}. In total, 56 distinct LMUs appear as the best match, with the most frequent accounting for only 20.9\% of perturbations. The median best-LMU overlap is seven genes, corresponding to a 19.9-fold enrichment over the random-overlap expectation. Fig.~\ref{fig:main_results}(d,e) shows representative LMU associations whose overlapping response genes form biologically coherent programs spanning interferon response, mitotic regulation, T-cell signaling and activation, oxidative phosphorylation, and RNA processing. LMUs are also largely non-redundant: across 50{,}000 randomly sampled pairs, the median candidate-gene Jaccard is $0.0079$, with 47.9\% sharing no genes and 80.1\% having Jaccard $<0.05$.

\section{Conclusion}
We introduced a mechanism-centric virtual cell world model that represents perturbation response as action-conditioned transitions over Latent Mechanism Units. Across held-out perturbations, VCLMU achieves strong response-gene recovery while maintaining global prediction accuracy, and the learned LMUs capture distinct and biologically coherent response programs. These results support mechanism-level latent state transition as a useful and interpretable formulation for virtual cell modeling. The current study focuses primarily on single-gene transcriptomic perturbations, and future work will extend LMU validation to multimodal and combinatorial perturbation settings.

\appendix
\newpage
\section{Additional Method Details}
\label{app:method_details}

\subsection{Notation and gene evidence}

We describe all quantities for a single sample $i$ consisting of a control reference and its associated perturbed profile. A superscript $c$ denotes the control state, a superscript $p$ denotes the perturbed state, and a hat denotes a model prediction. All losses below are per-sample; the training objective averages them over the mini-batch. The operator $\operatorname{Top}_{n}$ returns the indices of the $n$ largest scores, $\operatorname{norm}$ denotes $\ell_2$ normalization, $\operatorname{sg}$ denotes stop-gradient, and $\odot$ denotes elementwise product. All functions written $f_{\bullet}$ and all matrices written $W_{\bullet}$ are learned.

$\mathbf{x}_i^{c},\mathbf{x}_i^{p}\in\mathbb{R}^{G}$ are the control reference and perturbed profile, $a_i$ is the index of the perturbed gene, and $\Delta\mathbf{x}_i=\mathbf{x}_i^{p}-\mathbf{x}_i^{c}$. The model contains $K$ LMUs, of which a sparse subset $\mathcal{A}_i$ of size $K_a$ is active for sample $i$. Each LMU maintains $C$ candidate genes. Bold lower-case letters denote vectors; $\mathbf{H}_i^{c},\mathbf{H}_i^{p},\widehat{\mathbf{H}}_i^{p}\in\mathbb{R}^{K_a\times d}$ stack the corresponding LMU states. Table~\ref{tab:notation} summarizes the symbols used throughout.

\begin{table}[h]
\centering
\caption{Notation used in the main text and appendix.}
\label{tab:notation}
\begin{tabular}{ll}
\toprule
\textbf{Symbol} & \textbf{Meaning} \\
\midrule
$G$, $g$ & number of genes; gene index \\
$K$, $k$ & number of LMUs in the bank; LMU index \\
$K_a$, $\mathcal{A}_i$ & number of active LMUs for sample $i$; their index set \\
$C$, $\mathcal{C}_k$ & candidate genes per LMU; candidate set of LMU $k$ \\
$d$, $d_z$ & model width; latent transition width \\
$\mathbf{x}_i^{c}$, $\mathbf{x}_i^{p}$, $\widehat{\mathbf{x}}_i^{p}$ & control profile; observed perturbed profile; predicted perturbed profile \\
$\Delta\mathbf{x}_i$, $\widehat{\Delta\mathbf{x}}_i$ & observed response $\mathbf{x}_i^{p}-\mathbf{x}_i^{c}$; predicted response \\
$a_i$, $\boldsymbol{\psi}_i$ & perturbed gene (action); action embedding \\
$\mathcal{G}_L(a_i)$ & the $L$ highest-weight GRN neighbors of gene $a_i$ \\
$\mathbf{e}_g^{(m)}$, $\widetilde{\mathbf{e}}_g^{(m)}$, $\mathbf{e}_g$ & raw, projected, and fused multimodal evidence for gene $g$ \\
$\mathcal{M}_g$, $m$ & available modalities for gene $g$; modality index \\
$\mathbf{q}_k$, $\mathbf{q}_k^{0}$ & learnable LMU query; its initialization anchor \\
$\boldsymbol{\ell}_k$, $\bar{\boldsymbol{\ell}}_k$ & live LMU identity; EMA prototype \\
$\mathbf{o}_i^{c}$, $\mathbf{o}_i^{p}$ & observation summaries $f_{\mathrm{obs}}(\mathbf{x}_i^{c})$ and $f_{\mathrm{obs}}(\mathbf{x}_i^{p})$ \\
$w_{ik}^{c}$, $w_{ik}^{p}$ & LMU activity weights from $\mathbf{x}_i^{c}$ and $\mathbf{x}_i^{p}$ \\
$\mathbf{u}_k$, $\mathbf{b}_k$ & pooled candidate-gene features; identity--expression fusion \\
$\mathbf{h}_{ik}^{c}$, $\mathbf{h}_{ik}^{p}$, $\widehat{\mathbf{h}}_{ik}^{p}$ & control, observed perturbed, and predicted perturbed LMU state \\
$\mathbf{z}_{ik}$, $\boldsymbol{\mu}_{ik}$, $\boldsymbol{\sigma}_{ik}^{2}$ & transition latent; its conditional mean and variance \\
$\boldsymbol{\delta}_{ik}$, $\Delta\mathbf{h}_{ik}$ & raw and FiLM-modulated LMU state update \\
$\widehat{\mu}_{ig}$, $\widehat{v}_{ig}$ & predicted per-gene response mean and variance \\
$N_k$ & cumulative usage count of LMU $k$ \\
\bottomrule
\end{tabular}
\end{table}

Let $\mathbf{e}_g^{(m)}\in\mathbb{R}^{d_m}$ be the raw modality-$m$ feature for gene $g$, and let $\mathcal{M}_g$ be its set of available modalities. The evidence memory computes
\begin{equation}
\widetilde{\mathbf{e}}_g^{(m)}
=
\operatorname{LN}_m\!\left(W_m\mathbf{e}_g^{(m)}\right),
\qquad
\mathbf{e}_g
=
W_e
\left(
\frac{1}{|\mathcal{M}_g|}
\sum_{m\in \mathcal{M}_g}
\widetilde{\mathbf{e}}_g^{(m)}
\right).
\label{eq:app_evidence}
\end{equation}
All modality projections and $W_e$ are shared across genes. The implementation supports control-expression statistics, control co-expression PCs, Gene Ontology, Hallmark and PROGENy memberships, Gene2vec, STRING-PPI, Reactome, protein-language-model embeddings, and GTEx tissue expression. The evidence module supports per-gene modality masks; unavailable modalities are omitted from the average in Eq.~\eqref{eq:app_evidence}. Any evidence or GRN feature estimated from perturbational expression is fitted using the training split only.

The shared set encoder used below is a Transformer encoder followed by masked mean pooling and LayerNorm:
\begin{equation}
\operatorname{SetEnc}(\{\mathbf{t}_j\}_{j=1}^{n})
=
\operatorname{LN}
\left(
\frac{1}{n}\sum_{j=1}^{n}
\operatorname{Transformer}
(\{W_s\mathbf{t}_j\}_{j=1}^{n})_j
\right).
\label{eq:app_set_encoder}
\end{equation}

\subsection{LMU identities and candidate-gene refresh}

Each LMU has a learnable query $\mathbf{q}_k$ and an initialization anchor $\mathbf{q}_k^{0}$ obtained from gene-evidence centroids. The implementation permits deterministic tile means or $K$-means initialization; the reported training entry point uses deterministic tile means. Given candidate set $\mathcal{C}_k$, its live identity is
\begin{equation}
\boldsymbol{\ell}_k
=
\operatorname{LN}
\left(
\alpha\mathbf{q}_k+(1-\alpha)\operatorname{SetEnc}\{\mathbf{e}_g:g\in\mathcal{C}_k\}
\right),
\qquad
\alpha=\operatorname{sigmoid}(\alpha_0).
\label{eq:app_live_lmu}
\end{equation}
For selected LMUs, an exponential moving average (EMA) prototype is updated by
\begin{equation}
\bar{\boldsymbol{\ell}}_k
\leftarrow
c_{\mathrm{ema}}\bar{\boldsymbol{\ell}}_k
+
(1-c_{\mathrm{ema}})
\operatorname{sg}[\boldsymbol{\ell}_k].
\label{eq:app_ema}
\end{equation}
Candidate membership is refreshed periodically along the direction
\begin{equation}
\mathbf{n}_k
=
\operatorname{norm}
\left(
\tfrac{1}{2}\operatorname{norm}(\mathbf{q}_k)
+
\tfrac{1}{2}\operatorname{norm}(\bar{\boldsymbol{\ell}}_k)
\right),
\end{equation}
giving
\begin{equation}
\mathcal{C}_k
=
\operatorname{Top}_{C}
\left[
\cos(\mathbf{e}_g,\mathbf{n}_k)
+
c_{\mathrm{loy}}\,
\mathbf{1}\{g\in\mathcal{C}_k^{\mathrm{old}}\}
\right]_{g=1}^{G}.
\label{eq:app_candidate_refresh}
\end{equation}
The loyalty bonus $c_{\mathrm{loy}}$ provides hysteresis against abrupt membership replacement. Candidate sets may overlap; the model therefore does not impose a hard partition of genes. In distributed training, EMA statistics and usage counts are synchronized before every refresh.

\subsection{Control-based routing and state construction}

The action-independent observation encoder produces $\mathbf{o}_i^{c}=f_{\mathrm{obs}}(\mathbf{x}_i^{c})$ for the control reference and $\mathbf{o}_i^{p}=f_{\mathrm{obs}}(\mathbf{x}_i^{p})$ for the observed perturbed profile. Routing uses the control reference only:
\begin{align}
\mathcal{A}_i
&=
\operatorname{Top}_{K_a}
\left[
\cos\!\left(
\operatorname{norm}(W_o\mathbf{o}_i^{c}),
\operatorname{norm}(\bar{\boldsymbol{\ell}}_k)
\right)
+
\frac{c_{\mathrm{ucb}}}{\sqrt{N_k+1}}
\right]_{k=1}^{K},
\label{eq:app_hard_route}
\\
w_{ik}^{c}
&=
\frac{\exp\!\left(\cos(W_o\mathbf{o}_i^{c},\bar{\boldsymbol{\ell}}_k)/\tau\right)}
{\sum_{j\in\mathcal{A}_i}\exp\!\left(\cos(W_o\mathbf{o}_i^{c},\bar{\boldsymbol{\ell}}_j)/\tau\right)},
\quad k\in\mathcal{A}_i,
\label{eq:app_soft_route}
\end{align}
where $N_k$ is the cumulative usage count of LMU $k$ and
$\tau=0.01+0.99\operatorname{sigmoid}(\tau_0)$. The exploration bonus affects only the discrete top-$K_a$ selection; gradients flow through the soft activity weights. For the observed perturbed profile, $w_{ik}^{p}$ is recomputed from $\mathbf{o}_i^{p}$ over the fixed set $\mathcal{A}_i$.

Writing $\mathbf{y}$ for either profile, candidate-gene tokens, and expression-specific LMU features are
\begin{equation}
\mathbf{t}_{g}(\mathbf{y})
=
\mathbf{e}_g+\phi(y_g),
\qquad
\mathbf{u}_{k}(\mathbf{y})
=
\operatorname{SetEnc}
\{\mathbf{t}_{g}(\mathbf{y}):g\in\mathcal{C}_k\}.
\label{eq:app_expression_tokens}
\end{equation}
Identity and expression are fused before adding weak global context:
\begin{align}
\mathbf{b}_{k}(\mathbf{y})
&=
f_{\mathrm{fuse}}
\left(
[\boldsymbol{\ell}_k;\mathbf{u}_{k}(\mathbf{y})]
\right),
\\
\mathbf{h}_{k}(\mathbf{y})
&=
\rho_{k}(\mathbf{y})\,
f_{\mathrm{mix}}
\left(
[\mathbf{b}_{k}(\mathbf{y});
 \eta\,f_{\mathrm{obs}}(\mathbf{y})]
\right),
\label{eq:app_state_exact}\\
\rho_{k}(\mathbf{y})
&=
\operatorname{clip}
\left(
K_a\, w_{k}(\mathbf{y}),\,0.25,\,4
\right).
\end{align}
Together, $f_{\mathrm{fuse}}$, $f_{\mathrm{mix}}$, and the activity scaling $\rho_k$ constitute the state encoder $f_{\mathrm{state}}$ of Eq.~\eqref{eq:lmu_state}. The global-context gate is $\eta=\operatorname{sigmoid}(\eta_0)$ with $\eta_0$ initialized to $-3$, so the state initially depends primarily on LMU-specific candidate-gene expression. The same encoder and active indices define
$\mathbf{h}_{ik}^{c}=\mathbf{h}_{ik}(\mathbf{x}_i^{c})$ and
$\mathbf{h}_{ik}^{p}=\mathbf{h}_{ik}(\mathbf{x}_i^{p})$.

\subsection{Perturbation encoder and stochastic transition}

For the configured GRN-aware action encoder, let $\mathcal{G}_{L}(a_i)$ be the $L$ highest-weight GRN neighbors of perturbed gene $a_i$. The action representation is
\begin{equation}
\mathbf{c}_i
=
\operatorname{MHA}
\left(
\mathbf{e}_{a_i},
\{\mathbf{e}_{g}\}_{g\in\mathcal{G}_{L}(a_i)},
\{\mathbf{e}_{g}\}_{g\in\mathcal{G}_{L}(a_i)}
\right),
\qquad
\boldsymbol{\psi}_i
=
f_{\mathrm{act}}([\mathbf{e}_{a_i};\mathbf{c}_i]).
\label{eq:app_action}
\end{equation}
A simpler ablation uses $\boldsymbol{\psi}_i=\mathbf{e}_{a_i}$.

For each active LMU,
\begin{align}
\left[\boldsymbol{\mu}_{ik};\log\boldsymbol{\sigma}_{ik}^{2}\right]
&=
f_{\mathrm{prior}}
([\mathbf{h}_{ik}^{c};\boldsymbol{\psi}_i]),
\\
\mathbf{z}_{ik}
&=
\boldsymbol{\mu}_{ik}
+
\boldsymbol{\sigma}_{ik}
\odot\boldsymbol{\epsilon}_{ik},
\qquad
\boldsymbol{\epsilon}_{ik}\sim\mathcal{N}(\mathbf{0},\mathbf{I}),
\\
\boldsymbol{\delta}_{ik}
&=
f_{\mathrm{trans}}
([\mathbf{h}_{ik}^{c};\boldsymbol{\psi}_i;\mathbf{z}_{ik}]),
\\
[\boldsymbol{\gamma}_{ik};\boldsymbol{\beta}_{ik}]
&=
f_{\mathrm{film}}
([\mathbf{h}_{ik}^{c};\boldsymbol{\psi}_i]),
\\
\Delta\mathbf{h}_{ik}
&=
(1+\tanh\boldsymbol{\gamma}_{ik})
 \odot\boldsymbol{\delta}_{ik}
+
c_{\mathrm{res}}\tanh\boldsymbol{\beta}_{ik},
\qquad
\widehat{\mathbf{h}}_{ik}^{p}
&=
\mathbf{h}_{ik}^{c}+\Delta\mathbf{h}_{ik}.
\label{eq:app_dynamics_exact}
\end{align}
These blocks jointly implement $f_{\mathrm{dyn}}$ of Eq.~\eqref{eq:lmu_dynamics}. The implementation uses $c_{\mathrm{res}}=0.1$ to limit a deterministic additive bypass around $\mathbf{z}_{ik}$. During training, $\mathbf{z}_{ik}$ is sampled; at validation and inference, $\mathbf{z}_{ik}=\boldsymbol{\mu}_{ik}$. The KL term is
\begin{equation}
\mathcal{L}_{\mathrm{KL}}
=
\frac{1}{K_a}
\sum_{k\in\mathcal{A}_i}
\frac{1}{2}
\sum_{j=1}^{d_z}
\left[
\mu_{ikj}^{2}
+\sigma_{ikj}^{2}
-\log\sigma_{ikj}^{2}-1
\right].
\label{eq:app_kl}
\end{equation}
This is a KL-regularized conditional transition; there is no target-conditioned variational posterior.

\subsection{Observation decoder and inverse model}

A one-layer Transformer first permits interactions among predicted LMUs. Gene evidence is reused as decoder queries through $W_q$:
\begin{align}
\widetilde{\mathbf{H}}_i^{p}
&=
\operatorname{Transformer}_{\mathrm{LMU}}
(\widehat{\mathbf{H}}_i^{p}),
\\
\mathbf{d}_{ig}
&=
\operatorname{LN}
\left(
\operatorname{MHA}
(W_q\mathbf{e}_g,\widetilde{\mathbf{H}}_i^{p},\widetilde{\mathbf{H}}_i^{p})
+
W_q\mathbf{e}_g
\right),
\\
\left[\widehat{\mu}_{ig};\log\widehat{v}_{ig}\right]
&=
f_{\mathrm{gene}}([\mathbf{d}_{ig};W_q\mathbf{e}_g]).
\label{eq:app_decoder_exact}
\end{align}
Together these blocks form the decoder $f_{\mathrm{dec}}$ of Eq.~\eqref{eq:lmu_decoder}. The forward prediction loss is
\begin{equation}
\mathcal{L}_{\mathrm{fwd}}
=
\frac{1}{2G}
\sum_{g=1}^{G}
\left[
\log\widehat{v}_{ig}
+
\frac{(\Delta x_{ig}-\widehat{\mu}_{ig})^2}
{\widehat{v}_{ig}}
+
\log(2\pi)
\right].
\label{eq:app_nll}
\end{equation}

The inverse model predicts the perturbed gene using only the encoded state change. It forms per-LMU tokens
\begin{equation}
\mathbf{r}_{ik}
=
f_{\mathrm{inv}}
\left(
[\mathbf{h}_{ik}^{c};
 \mathbf{h}_{ik}^{p};
 \mathbf{h}_{ik}^{p}-\mathbf{h}_{ik}^{c};
 \mathbf{h}_{ik}^{p}\odot\mathbf{h}_{ik}^{c}]
\right),
\end{equation}
attention-pools them into $\bar{\mathbf{r}}_i$, and computes
\begin{equation}
s_{ig}
=
\exp(\xi)
\cos\!\left(
\bar{\mathbf{r}}_i,
W_{\mathrm{inv}}\mathbf{e}_g
\right),
\qquad
\mathcal{L}_{\mathrm{inv}}
=
\operatorname{CE}
(\mathbf{s}_i,a_i),
\label{eq:app_inverse}
\end{equation}
where $\xi$ is a learnable logit scale and $\mathbf{s}_i=(s_{i1},\ldots,s_{iG})$. The true perturbation identity $a_i$ is used only as the classification target, not as an input to the inverse head.

\subsection{Auxiliary supervision and LMU regularization}

Because the active indices are shared, dynamics supervision is position-wise:
\begin{equation}
\mathcal{L}_{\mathrm{dyn}}
=
\operatorname{SmoothL1}
\left(
\widehat{\mathbf{H}}_i^{p}-\mathbf{H}_i^{c},
\operatorname{sg}
[\mathbf{H}_i^{p}-\mathbf{H}_i^{c}]
\right).
\label{eq:app_transition}
\end{equation}
The stop-gradient prevents the target branch from moving merely to match the prediction. To ground that branch independently,
\begin{equation}
\bar{\mathbf{h}}_i^{p}
=
\sum_{k\in\mathcal{A}_i}
w_{ik}^{p}\mathbf{h}_{ik}^{p},
\qquad
\mathcal{L}_{\mathrm{obs}}
=
\operatorname{SmoothL1}
\left(
W_{\mathrm{out}}(\bar{\mathbf{h}}_i^{p}),
\mathbf{x}_i^{p}
\right).
\label{eq:app_state_grounding}
\end{equation}

The structural regularizer is
\begin{equation}
\mathcal{R}_{\mathrm{LMU}}
=
\lambda_{\mathrm{div}}\mathcal{R}_{\mathrm{div}}
+
\lambda_{\mathrm{anc}}\mathcal{R}_{\mathrm{anc}}
+
\lambda_{\mathrm{act}}\mathcal{R}_{\mathrm{act}},
\label{eq:app_lmu_reg}
\end{equation}
with
\begin{align}
\mathcal{R}_{\mathrm{div}}
&=
\mathbb{E}_{k\neq l}
\left[
\max\{0,\cos(\boldsymbol{\ell}_k,\boldsymbol{\ell}_l)-c_{\mathrm{div}}\}^{2}
\right],
\\
\mathcal{R}_{\mathrm{anc}}
&=
\frac{1}{K}
\sum_{k=1}^{K}
\left[
1-\cos(\mathbf{q}_k,\mathbf{q}_k^{0})
\right],
\\
H_i
&=
-\sum_{k\in\mathcal{A}_i}
 w_{ik}^{c}\log(w_{ik}^{c}+10^{-8}),
\qquad
\mathcal{R}_{\mathrm{act}}
=
\left(
\mathbb{E}_{i}[H_i]
-H_{\star}
\right)^2,
\label{eq:app_structural_terms}
\end{align}
where $\mathbb{E}_{i}[H_i]$ is the mini-batch mean of the per-cell activity entropy. The diversity expectation is estimated from sampled pairs of live, active LMUs. Unlike plain entropy minimization, $\mathcal{R}_{\mathrm{act}}$ targets a moderate entropy $H_{\star}$ and therefore discourages both one-LMU collapse and uniform activity.

Combining the terms gives
\begin{equation}
\mathcal{L}
=
\mathcal{L}_{\mathrm{fwd}}
+\lambda_{\mathrm{dyn}}\mathcal{L}_{\mathrm{dyn}}
+\lambda_{\mathrm{obs}}\mathcal{L}_{\mathrm{obs}}
+\lambda_{\mathrm{KL}}\mathcal{L}_{\mathrm{KL}}
+\lambda_{\mathrm{inv}}\mathcal{L}_{\mathrm{inv}}
+\mathcal{R}_{\mathrm{LMU}},
\label{eq:app_total}
\end{equation}
where $\lambda_{\mathrm{KL}}$ and, when used, $\lambda_{\mathrm{inv}}$ follow linear warm-up schedules that scale them from $0$ to their final values, and all per-sample terms are averaged over the mini-batch.

\subsection{Two-stage training and inference}

We first form perturbation-disjoint data splits before any response-derived preprocessing or model fitting. All response-derived preprocessing, control statistics, GRN construction, and evidence fitting are performed using training data only.

\paragraph{Stage 1: pseudo-bulk pretraining.}
For each perturbed pseudo-bulk profile, $\mathbf{x}_i^{c}$ is the training control mean for its context. The model is optimized end-to-end, including evidence projections, set encoder, LMU queries, dynamics, and decoder. Candidate memberships are refreshed periodically from synchronized EMA prototypes. The current configuration jointly uses forward prediction, dynamics alignment, target-state grounding, inverse prediction, KL regularization, and LMU regularization.

\paragraph{Stage 2: single-cell pretraining.}
The stage-1 checkpoint is restored before recomputing evidence-derived quantities. Single-cell profiles are log-transformed and aligned one-to-one to the stage-1 gene order; only shared genes contribute to observation losses. The control reference is estimated from training control cells. Evidence projections are then frozen, and the remaining model continues pretraining at a lower learning rate. In the current configuration, the dynamics and target-state-grounding terms are disabled, while the forward and inverse objectives remain active.

At inference, the observed perturbed profile is absent: the model selects LMUs from $\mathbf{x}_i^{c}$, encodes $a_i$, sets $\mathbf{z}_{ik}=\boldsymbol{\mu}_{ik}$, predicts $\widehat{\mathbf{H}}_i^{p}$, and returns
$\widehat{\mathbf{x}}_i^{p}=\mathbf{x}_i^{c}+\widehat{\boldsymbol{\mu}}_i$. The target-state, state-grounding, and inverse branches are not required.

\section{Data}
\label{sec:data}

\paragraph{Pseudo-bulk corpus.}
Stage-1 training uses approximately 200K pseudo-bulk genetic perturbation profiles constructed from publicly available datasets comprising more than 10 million single-cell perturbation profiles, including X-Atlas-Orion~\cite{xatlasorion2025}, Marson~\cite{marson2025}, and the Virtual Cell Challenge 2025 collection~\cite{roohani2025vcc}. We use only gene perturbation experiments (e.g., CRISPR knockout, CRISPR interference, and CRISPR activation) and exclude chemical perturbation data. Pseudo-bulk profiles are generated by randomly aggregating 200 cells from the same perturbation condition within each cellular context, improving the signal-to-noise ratio while preserving perturbation-specific and context-specific biological responses. Expression profiles are normalized and log-transformed over a shared gene vocabulary. For each perturbed pseudo-bulk profile, the control reference $\mathbf{x}_i^{c}$ is defined as the mean expression profile of the corresponding training control cells from the same cellular context.

\paragraph{Single-cell data.}
Stage 2 continues pretraining on gene-aligned single-cell genetic perturbation data derived from the same resources, including X-Atlas-Orion~\cite{xatlasorion2025}, Marson~\cite{marson2025}, and the Virtual Cell Challenge 2025 collection~\cite{roohani2025vcc}. Profiles are library-size normalized and log-transformed. Only genes shared between Stage 1 and Stage 2 contribute to the observation losses. For each dataset, the control reference is estimated as the mean expression profile of the corresponding training control cells.

\paragraph{Evaluation Datasets}
 Evaluation datasets are Adamson~\citep{adamson2016}, Norman~\citep{norman2019}, Replogle K562 and RPE1~\citep{replogle2022}, and Tian CRISPRa and CRISPRi~\citep{tian2021}, spanning multiple genetic perturbation screens and cellular contexts.

\paragraph{Gene evidence.}
Each gene carries a multimodal evidence embedding built from co-expression, Gene Ontology, MSigDB Hallmark, PROGENy, Gene2Vec, STRING Mashup, Reactome, ESM2, and GTEx representations (Appendix~\ref{app:evidence}). Features estimated from perturbational expression are fitted using the training split only.

\section{Biological Evidence Representations}
\label{app:evidence}

To incorporate complementary biological evidence, we augment each gene with a collection of representations capturing transcriptional co-regulation, functional annotation, signaling pathways, interaction networks, protein sequence, and tissue-specific expression. These biological evidence representations combine information from curated biological resources with training-derived expression statistics. All representations are constructed at the human gene level, aligned to the shared gene vocabulary used throughout training, and standardized prior to model training. Each evidence modality $m$ provides the raw feature $\mathbf{e}_g^{(m)}$ used in Eq.~\eqref{eq:app_evidence}.

\begin{table*}[h]
\centering
\small
\caption{Biological evidence representations used to construct the multimodal gene evidence embedding.}
\label{tab:bio_evidence}
\setlength{\tabcolsep}{4pt}
\begin{tabularx}{\textwidth}{
>{\centering\arraybackslash}p{0.7cm}
>{\raggedright\arraybackslash}p{3.5cm}
>{\raggedright\arraybackslash}X
>{\centering\arraybackslash}p{0.8cm}
}
\toprule
\textbf{Evidence} & \textbf{Source} & \textbf{Description} & \textbf{Dim.} \\
\midrule

$m_1$ & Training co-expression &
Principal components of the training-set gene--gene co-expression matrix, capturing major transcriptional covariance patterns. &
$\leq256$ \\

$m_2$ & Gene Ontology~\cite{go2021} &
Gene--term association matrix compressed using truncated SVD to summarize biological process, molecular function, and cellular component annotations. &
512 \\

$m_3$ & MSigDB Hallmark~\cite{liberzon2015hallmark} &
Binary membership in the Hallmark gene sets, representing approximately 50 curated biological programs. &
$\sim50$ \\

$m_4$ & PROGENy~\cite{schubert2018progeny} &
Association with the 14 PROGENy signaling pathways, providing pathway-level functional information. &
14 \\

$m_5$ & Gene2Vec~\cite{gene2vec2018} &
Pretrained gene embeddings capturing functional similarity learned from large-scale biological co-occurrence. &
200 \\

$m_6$ & STRING + Mashup~\cite{string2023,mashup2016} &
Protein interaction network embedding capturing both local and global network topology. &
800 \\

$m_7$ & Reactome~\cite{reactome2024} &
Gene--pathway association matrix compressed using truncated SVD to summarize Reactome pathway memberships. &
200 \\

$m_8$ & ESM2-3B~\cite{esm2} &
Protein language model embedding encoding evolutionary and structural information from protein sequences. &
2560 \\

$m_9$ & GTEx v8~\cite{gtex2020} &
RNA-seq expression profiles across 54 human tissues, providing tissue-specific biological context. &
54 \\

\bottomrule
\end{tabularx}
\end{table*}

Among these evidence representations, only the co-expression representation ($m_1$) is estimated from the training data. All remaining representations are obtained from external biological resources. PCA is used to construct the co-expression representation, while truncated SVD is applied to the Gene Ontology and Reactome annotation matrices to obtain compact embeddings.

\section{Experimental Details}
\label{sec:exp}

\subsection{Perturbation-disjoint Data Splits}
\label{sec:data_split}
The perturbation-disjoint constraint is applied to downstream fine-tuning and evaluation splits. Because Stage~1 and Stage~2 pretraining use separate large-scale perturbation corpora, a downstream test perturbation may still appear during pretraining in a different dataset or cellular context.
Following the perturbation-disjoint evaluation setting used in prior perturbation-response benchmarks, we split each evaluation dataset by perturbation identity rather than by individual cells. For each random seed, 25\% of the genetic perturbations are held out for testing, and all cells associated with a held-out perturbation are assigned to the test set. The remaining perturbations are used for downstream model development, ensuring that no test perturbation appears as an intervention during downstream model fitting. This setting evaluates generalization to held-out perturbation identities rather than to new cells from perturbations already observed during training. We repeat the perturbation-level split across three random seeds and report performance aggregated over the resulting test sets. For downstream evaluation, VCLMU-FT initializes from the Stage-2 pretrained checkpoint and is further fine-tuned using only the downstream training split of each evaluation dataset. Held-out perturbations are never used during downstream fine-tuning.

\subsection{Baseline Implementation Details}
\label{app:baselines}

All baselines are evaluated under the same perturbation-disjoint splits described in Appendix~\ref{sec:data_split}. For each dataset and random seed, models are trained using only the downstream training split and evaluated on held-out perturbation identities. For each held-out perturbation, we compare the observed condition-level mean expression profile with the corresponding mean predicted profile.

\paragraph{GEARS.}
We use the official \texttt{cell-gears} implementation~\citep{roohani2024gears}. GEARS propagates perturbation effects over a gene graph constructed from biological relationships. The model is trained independently for each dataset and random split with hidden size $64$, batch size $32$, and $50$ training epochs using Adam with learning rate $10^{-3}$. Predictions for each held-out perturbation are obtained using the model's standard perturbation-conditioned prediction procedure.

\paragraph{CPA.}
We use the \texttt{cpa-tools} implementation~\citep{lotfollahi2023cpa}. The model uses a $64$-dimensional latent representation with a Gaussian reconstruction objective. The encoder contains four layers of width $256$ with batch normalization and dropout $0.2$, while the decoder contains two layers of width $256$. Perturbation identity, dosage, and cell type are represented as covariates, together with the adversarial training objective of CPA. Models are trained with Adam at learning rate $10^{-3}$ for up to $50$ epochs with early stopping. At inference, control cells are provided together with the target perturbation, and predicted profiles are averaged to obtain the condition-level prediction.

\paragraph{scGPT and scGPT-FT.}
We use the \texttt{TransformerGenerator} architecture from scGPT~\citep{cui2024scgpt}, with embedding dimension $512$, feed-forward dimension $512$, $12$ Transformer layers, $8$ attention heads, and dropout $0.2$. Perturbation identity is represented as a binary per-gene perturbation indicator together with gene expression values. \textbf{scGPT} is trained from random initialization, whereas \textbf{scGPT-FT} initializes the encoder, value encoder, and Transformer encoder from the pretrained \texttt{scGPT\_human} checkpoint before downstream fine-tuning. Both variants are trained with batch size $8$ using Adam with learning rate $10^{-4}$ for up to $15$ epochs and early stopping on validation masked MSE. At inference, predictions are generated from sampled control cells paired with the target perturbation and averaged to obtain the condition-level prediction.

\paragraph{Perturbed Mean.}
We include a condition-agnostic reference that predicts the same perturbation response for every held-out perturbation. Specifically, the prediction is the mean expression profile pooled over all non-control cells in the downstream training split, without using the identity of the test perturbation.

\subsection{LMU Interpretability Analysis}
\label{app:lmu_interpretability}

We evaluate the interpretability of the learned LMUs from two complementary perspectives: whether different LMUs form non-redundant gene modules, and whether these modules align with perturbation-specific transcriptional responses. Each background LMU $k$ is associated with a learned candidate gene set $\mathcal{C}_k$, defined by the candidate gene indices retained by the trained model.

\paragraph{LMU non-redundancy.}
We first assess whether the learned LMU library collapses to highly overlapping gene modules. For two LMUs $k$ and $l$, we compute the Jaccard similarity between their candidate gene sets,
\[
J(k,l)
=
\frac{
|\mathcal{C}_k \cap \mathcal{C}_l|
}{
|\mathcal{C}_k \cup \mathcal{C}_l|
}.
\]
We evaluate the distribution of Jaccard similarities over 50{,}000 randomly sampled pairs of distinct LMUs. We report the median Jaccard similarity, the fraction of sampled pairs with zero candidate-gene overlap, and the fraction with Jaccard similarity below $0.05$. These statistics quantify whether the learned LMU library contains distinct rather than highly redundant gene modules.

\paragraph{Perturbation-specific response construction.}
We next test whether held-out perturbations preferentially align with different learned LMUs. For each perturbation $a_i$, we compute an observed transcriptional response relative to its matched control context. For every observation associated with $a_i$, the corresponding context-specific control reference is subtracted from the perturbed expression profile, and the resulting response vectors are averaged over observations carrying the same perturbation. We denote the resulting mean response by $\Delta\mathbf{x}_i$. Only held-out perturbations represented by at least 10 observations are retained for this analysis.

Because raw perturbation responses can contain transcriptional changes shared across many perturbations, we explicitly remove ubiquitous and shared response signals before measuring LMU alignment. We first identify genes that recurrently appear among strong raw responses. For each retained perturbation, we select the 200 genes with the largest absolute values in $\Delta\mathbf{x}_i$, excluding the directly perturbed gene itself. Let $\mathcal{T}_i^{\mathrm{raw}}$ denote this set. For each gene $g$, we compute its prevalence across perturbations,
\[
\pi_g
=
\frac{1}{N}
\sum_{i=1}^{N}
\mathbf{1}
\left[
g \in \mathcal{T}_i^{\mathrm{raw}}
\right],
\]
where $N$ is the number of retained perturbations. Genes with $\pi_g>0.20$ are treated as ubiquitous response genes and excluded from subsequent enrichment analysis.

We additionally subtract the signed response shared across perturbations. Specifically, we compute the gene-wise median response
\[
\mathbf{m}
=
\operatorname{median}_{i}
\left(
\Delta\mathbf{x}_i
\right),
\]
and define the residual perturbation-specific response as
\[
\mathbf{r}_i
=
\Delta\mathbf{x}_i-\mathbf{m}.
\]
Among genes not classified as ubiquitous, we select the 100 genes with the largest absolute residual response $|\mathbf{r}_i|$ for each perturbation, again excluding the directly perturbed gene. We denote this perturbation-specific response set by $\mathcal{S}_i$.

\paragraph{Perturbation--LMU enrichment.}
For each perturbation--LMU pair $(i,k)$, we quantify whether the perturbation-specific response set $\mathcal{S}_i$ significantly overlaps the candidate gene set of LMU $k$. Ubiquitous response genes are removed from $\mathcal{C}_k$ before testing. Let $\widetilde{\mathcal{C}}_k$ denote the resulting LMU candidate set and let $G_{\mathrm{eff}}$ denote the number of genes remaining in the effective background universe.

We compute the observed overlap
\[
O_{ik}
=
|\mathcal{S}_i \cap \widetilde{\mathcal{C}}_k|,
\]
and evaluate its significance using a one-sided hypergeometric test with population size $G_{\mathrm{eff}}$, $|\widetilde{\mathcal{C}}_k|$ candidate genes, and $|\mathcal{S}_i|$ sampled response genes. The expected overlap under the corresponding random-overlap null is
\[
E_{ik}
=
\frac{
|\mathcal{S}_i|
|\widetilde{\mathcal{C}}_k|
}{
G_{\mathrm{eff}}
}.
\]
We define fold enrichment as
\[
F_{ik}
=
\frac{O_{ik}}{E_{ik}}.
\]

For each perturbation independently, the hypergeometric $p$-values obtained across all LMUs are corrected using the Benjamini--Hochberg procedure~\citep{benjamini1995fdr}. We consider a perturbation--LMU association significant when the resulting FDR is below $0.05$. For visualization, association strength is represented as
\[
-\log_{10}(\mathrm{FDR}_{ik}).
\]
The best-matching LMU for perturbation $i$ is defined as the LMU with the largest $-\log_{10}(\mathrm{FDR}_{ik})$.

Summary statistics are computed over all retained held-out perturbations. We report the fraction of perturbations whose best-matching LMU has $\mathrm{FDR}<0.05$, the number of distinct LMUs appearing as a best match, and the fraction of perturbations assigned to the most frequently selected best-matching LMU. We additionally report the median observed overlap $O_{ik}$ and the median fold enrichment $F_{ik}$ evaluated at the best-matching LMU of each perturbation. These quantities respectively summarize the prevalence, diversity, concentration, overlap size, and enrichment strength of perturbation-specific LMU associations.

\paragraph{Representative biological programs.}
To provide qualitative examples of the biological structure captured by the learned units, we further examine representative significant perturbation--LMU associations. For each selected pair, we inspect the genes in the intersection
\[
\mathcal{S}_i \cap \widetilde{\mathcal{C}}_k,
\]
which are simultaneously among the strongest perturbation-specific response genes and members of the corresponding learned LMU candidate set.

The six representative associations shown in Fig.~\ref{fig:main_results} are EDRF1--LMU~905, SLC10A7--LMU~332, ANKRD16--LMU~325, TMEM33--LMU~925, CARMIL1--LMU~439, and SCML2--LMU~1836. Biological program labels are assigned based on the functional coherence of the overlapping genes rather than through an automated pathway-enrichment procedure. For example, the EDRF1--LMU~905 overlap contains interferon-associated genes including \textit{IFI44}, \textit{IFI44L}, \textit{IFI6}, \textit{IFIH1}, \textit{IFIT3}, \textit{ISG15}, \textit{MX1}, \textit{OAS1}, \textit{OAS2}, and \textit{RSAD2}, motivating its annotation as an interferon-response program. Similarly, the other representative overlaps contain coherent gene groups related to mitotic regulation, T-cell signaling, T-cell activation, oxidative phosphorylation, and RNA processing.

\subsection{Evaluation Metrics}
\label{sec:evaluation_metrics}

We evaluate perturbation-response predictions independently for each held-out perturbation condition. For a test perturbation $p$, let $\mathbf{x}_{p}\in\mathbb{R}^{G}$ denote the observed condition-level expression profile and $\widehat{\mathbf{x}}_{p}\in\mathbb{R}^{G}$ denote the corresponding model prediction over the same $G$ genes. All control reference profiles are constructed exclusively from the training split, while ground-truth differential-expression sets are used only for evaluation. We report three complementary metrics: $\Delta$PCC for global response prediction, $\Delta$PCC@20 for prediction of strongly responsive genes, and Jaccard@20 for recovery of response-gene identities.

\paragraph{Control-referenced perturbation response.}
We first construct a common control reference using all control cells in the training split. Let $\mathcal{X}_{\mathrm{ctrl}}$ denote the set of training control profiles. The training control mean is
\begin{equation}
\bar{\mathbf{x}}_{\mathrm{ctrl}}
=
\frac{1}{|\mathcal{X}_{\mathrm{ctrl}}|}
\sum_{\mathbf{x}\in\mathcal{X}_{\mathrm{ctrl}}}
\mathbf{x}.
\label{eq:eval_control_mean}
\end{equation}
For each held-out perturbation $p$, the observed and predicted perturbation-induced expression changes are defined relative to the same control reference:
\begin{equation}
\Delta\mathbf{x}_{p}
=
\mathbf{x}_{p}
-
\bar{\mathbf{x}}_{\mathrm{ctrl}},
\qquad
\widehat{\Delta\mathbf{x}}_{p}
=
\widehat{\mathbf{x}}_{p}
-
\bar{\mathbf{x}}_{\mathrm{ctrl}}.
\label{eq:eval_delta}
\end{equation}
Using the same training-derived control reference for both observed and predicted profiles ensures that the evaluation measures agreement in the perturbation-induced response under a common baseline.

\paragraph{Global perturbation-response correlation.}
We use Pearson correlation across all evaluated genes to measure agreement between the observed and predicted perturbation-response profiles:
\begin{equation}
\Delta\mathrm{PCC}_{p}
=
\operatorname{PCC}
\left(
\Delta\mathbf{x}_{p},
\widehat{\Delta\mathbf{x}}_{p}
\right).
\label{eq:eval_delta_pcc}
\end{equation}
This metric evaluates whether the model reproduces the direction and relative magnitude of gene-expression changes induced by a held-out perturbation across the full transcriptional response.

\paragraph{DE-focused perturbation-response correlation.}
Because the global correlation includes many genes that change only weakly under a perturbation, we additionally evaluate prediction accuracy on the most strongly responsive genes. For each perturbation $p$, we use the benchmark-provided set of the top 20 non-dropout differentially expressed genes identified from the observed perturbation response, denoted by
\begin{equation}
\mathcal{D}_{p}
=
\{g_1,\ldots,g_{20}\},
\qquad
|\mathcal{D}_{p}|=20 .
\end{equation}
The set $\mathcal{D}_{p}$ is defined exclusively from the ground-truth observations and is independent of model predictions. We then compute Pearson correlation after restricting both response vectors to these genes:
\begin{equation}
\Delta\mathrm{PCC@20}_{p}
=
\operatorname{PCC}
\left(
\Delta\mathbf{x}_{p}[\mathcal{D}_{p}],
\widehat{\Delta\mathbf{x}}_{p}[\mathcal{D}_{p}]
\right).
\label{eq:eval_delta_pcc20}
\end{equation}
While $\Delta$PCC measures agreement over the complete transcriptional response, $\Delta$PCC@20 focuses specifically on whether the model correctly captures the direction and relative magnitude of expression changes among the genes most strongly affected by the perturbation.

\paragraph{Top-20 response-gene recovery.}
$\Delta$PCC@20 evaluates prediction accuracy on genes that are already known from the ground-truth experiment to be highly responsive, but it does not measure whether the model itself can identify which genes should respond most strongly. We therefore additionally report Jaccard@20.

Genes are ranked by the absolute magnitude of their predicted perturbation-induced change relative to the training control mean, giving the predicted top-20 response set
\begin{equation}
\widehat{\mathcal{D}}_{p}
=
\operatorname{Top}_{20}
\left\{
\left|
\widehat{\Delta x}_{pg}
\right|
\right\}_{g=1}^{G}.
\label{eq:eval_pred_top20}
\end{equation}
We then compare the predicted response genes with the ground-truth top-20 non-dropout DE genes using the Jaccard index:
\begin{equation}
\operatorname{Jaccard@20}_{p}
=
\frac{
\left|
\mathcal{D}_{p}
\cap
\widehat{\mathcal{D}}_{p}
\right|
}{
\left|
\mathcal{D}_{p}
\cup
\widehat{\mathcal{D}}_{p}
\right|
}.
\label{eq:eval_jaccard20}
\end{equation}
Jaccard@20 is therefore complementary to $\Delta$PCC@20: the latter measures expression agreement on the experimentally identified responsive genes, whereas Jaccard@20 evaluates whether the model can recover the identities of those genes from its own predicted response.

\paragraph{Aggregation.}
All three metrics are computed independently for each held-out perturbation condition. Condition-level scores are then averaged across the held-out perturbations within each split, and the evaluation is repeated over the three perturbation-disjoint random splits described above. We report the resulting mean performance across the three splits, with variability summarized by the corresponding standard deviation.




\end{document}